\documentclass[letterpaper, 10 pt, conference]{ieeeconf}  

\IEEEoverridecommandlockouts                              

\usepackage{graphics} 
\usepackage{epsfig} 
\usepackage{times} 
\usepackage{amssymb}  
\usepackage{xcolor}
\usepackage{color, colortbl}
\usepackage{booktabs}
\usepackage{amsmath}

\newcommand{\handy}[1]{\textcolor{brown}{#1}}

\title{\LARGE \bf
OmniColor: A Global Camera Pose Optimization Approach of LiDAR-360Camera Fusion for Colorizing Point Clouds
}

\author{Bonan Liu$^{1*}$, Guoyang Zhao$^{1*}$, Jianhao Jiao$^{1}$, Guang Cai$^{2}$, Chengyang Li$^{1}$, \\Handi Yin$^{1}$, Yuyang Wang$^{1}$, Ming Liu$^{1,2}$ and Pan Hui$^{1,2}$
\vspace{-10pt}
\thanks{*\textbf{Indicates Equal Contribution.}}
\thanks{$^{1}$B. Liu, G. Zhao, J. Jiao, C. Li, H. Yin, Y. Wang, IEEE Member, M. Liu, IEEE Senior Member, P. Hui, IEEE Fellow, are with The Hong Kong University of Science and Technology (Guangzhou), Guangzhou, China}%
\thanks{$^{2}$G. Cai is with The Hong Kong University of Science and Technology, Hong Kong SAR, China}%
\thanks{E-mail: \{bliu404, gzhao492, cli386, hyin335\}@connect.hkust-gz.edu.cn, jiaojh1994@gmail.com, gcaiab@connect.ust.hk, {\{yuyangwang, eelium, panhui}\}@ust.hk}}

\begin{document}
\maketitle
\thispagestyle{empty}
\pagestyle{empty}

\begin{abstract}
A Colored point cloud, as a simple and efficient 3D representation, has many advantages in various fields, including robotic navigation and scene reconstruction. 
This representation is now commonly used in 3D reconstruction tasks relying on cameras and LiDARs. However, fusing data from these two types of sensors is poorly performed in many existing frameworks, leading to unsatisfactory mapping results, mainly due to inaccurate camera poses. This paper presents OmniColor, a novel and efficient algorithm to colorize point clouds using an independent 360-degree camera. Given a LiDAR-based point cloud and a sequence of panorama images with initial coarse camera poses, our objective is to jointly optimize the poses of all frames for mapping images onto geometric reconstructions. Our pipeline works in an off-the-shelf manner that does not require any feature extraction or matching process. Instead, we find optimal poses by directly maximizing the photometric consistency of LiDAR maps. In experiments, we show that our method can overcome the severe visual distortion of omnidirectional images and greatly benefit from the wide field of view (FOV) of 360-degree cameras to reconstruct various scenarios with accuracy and stability. The code will be released at https://github.com/liubonan123/OmniColor/.

\end{abstract}

\section{INTRODUCTION}
Over the last two decades, various reality capture methods have been successfully designed for reconstructing large-scale environments by using cameras, LiDAR, as well as other perceptual sensors. Camera-based photogrammetry methods \cite{nocerino20173d,hussain2022experimental} typically extract visual features from textured scenes and then utilize the overlapping areas to align photos taken from different angles. 
However, these methods are highly susceptible to variations in illumination and visual complexity of scenes.
LiDAR-based scanning methods \cite{zhang2014loam,shan2018lego}, which typically extract structural features from point clouds, are invariant to such changes.
However, a fundamental drawback of LiDAR is that they do not provide rich visual appearance information, which is exceptionally beneficial for humans to recognize geometric information from enormous amounts of point clouds.
Recently, some systems \cite{vechersky2018colourising,lin2022r,lin2022rr,zheng2022fast,zhen2019joint} have been designed to utilize the complementary properties of LiDARs and cameras to create colored point cloud maps but still suffer from blurring, ghosting, and other visual artifacts, due to inaccurate camera poses and illumination variations.
\handy{
}

To mitigate the problems above, an effective strategy is to enlarge the camera field of view (FOV) \cite{sumikura2019openvslam}. The expansion of FOV facilitates capturing additional scene information, broadening the co-visible regions among successive frames. 
It enables the system to capture the surrounding area simultaneously to reduce the artifacts from illumination variation and acquire sufficient correspondence for improved adaptive ability in diverse environments. Concurrently, progress in lens manufacturing techniques has made it possible to attain complete 360-degree perception with only two lenses mounted back-to-back \cite{huang2022360vo}, 
providing a cost-effective mobile mapping solution.
\begin{figure}[t]
    \centering
    \includegraphics[width=0.49\textwidth]{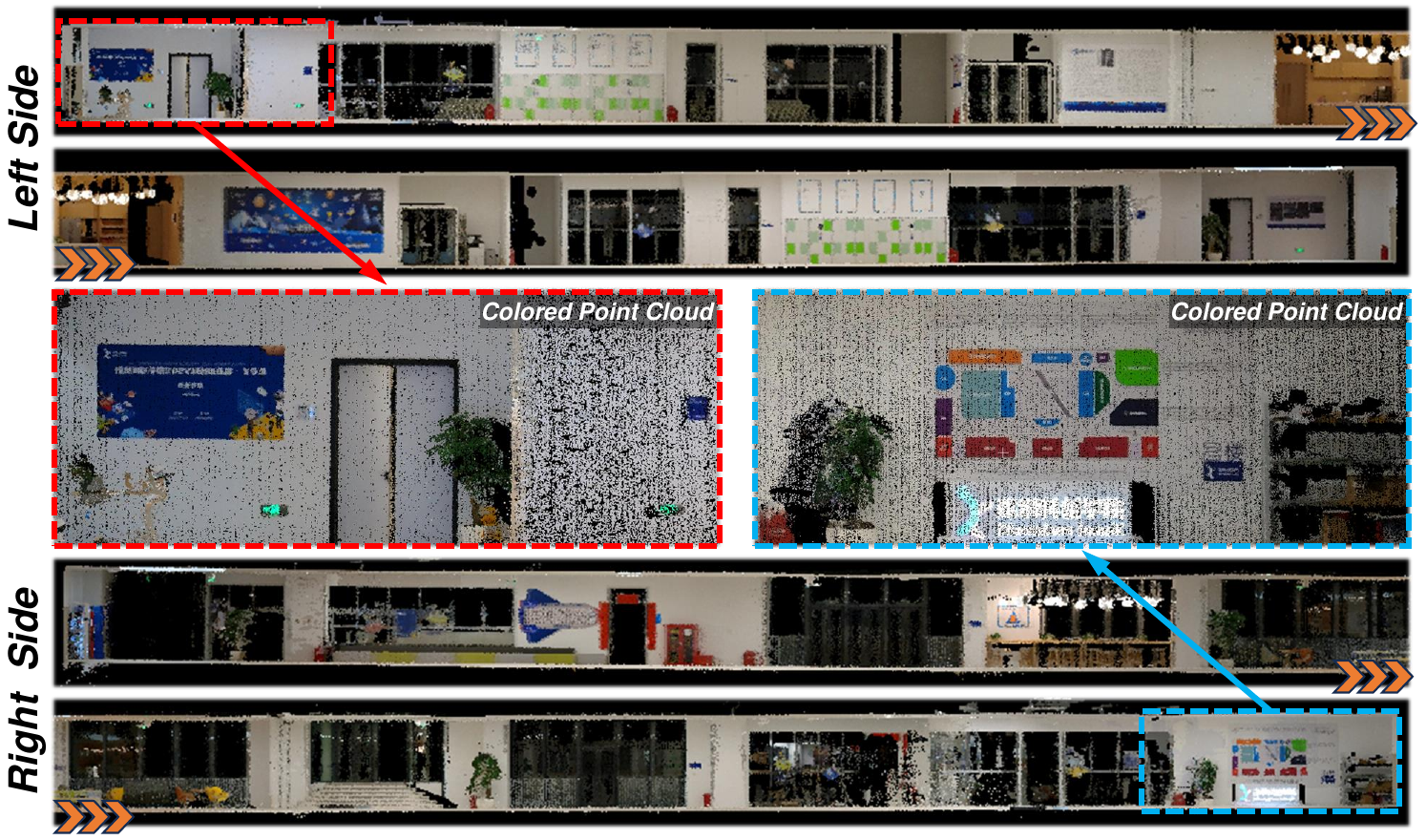}
    \vspace{-18pt}
    \caption{\small{The RGB-Colored point cloud reconstructed by our mobile mapping system. A fly-through rendered image from the point cloud's left and right sides.}}
    \label{coverfigure}
    \vspace{-17pt}
\end{figure}

For customized handheld mobile mapping devices with an extra commercial 360-degree camera \cite{cui2022cp+}, several issues may negatively affect the quality of reconstructed colored point clouds. First, the input geometric model \cite{xu2022fast} for which the colored point cloud is constructed is inaccurate and produced from noisy data. 
Second, LiDARs and cameras are not hardware-synchronized. The color images are not in rigid correspondence with the LiDAR range measurements. Third, panorama images are subject to significant visual distortion caused by the spherical projection process. 
Lastly, establishing correct correspondences among data 
in heterogeneous modalities (\textit{e.g.,} geometry and texture) is challenging.
For instance, several systems \cite{yuan2021pixel,miao2023coarse} that rely on 2D-3D geometric line correspondences often encounter misalignment issues between images and LiDAR-reconstructed maps.

In this paper, we describe an approach for optimizing the mapping
of color images produced by an independent 360-degree camera to a corresponding geometric reconstruction. Our method addresses the above problems through a cohesive optimization framework, wherein we jointly optimize the poses of all color frames to rectify inaccurate colorization regions. The camera poses for all images are globally optimized to maximize a unified objective: the photometric consistency of the colored point cloud maps. Furthermore, our approach is specially adjusted and highly suitable for 360-degree cameras. It allows us to bypass the non-differentiable changes in the visibility relationship between the camera and point cloud during optimization, significantly reducing the computational cost. 
Overall, our contributions can be summarized as follows:

\begin{itemize}
\item[$\bullet$] We propose a novel global optimization approach of LiDAR-360 camera fusion for convenient and precise point cloud colorization (see Fig.~\ref{coverfigure}), which can overcome the severe visual distortion in omnidirectional images and gain benefit from a wider FOV.
\item[$\bullet$] We propose a novel point cloud co-visibility estimation approach, which mitigates the impact of the noise in point cloud surfaces on visibility relationships.
\item[$\bullet$] Our approach operates in a readily available manner, enabling seamless integration with any mobile mapping system while ensuring both convenience and accuracy. Extensive experiments demonstrate its superiority over existing frameworks. 
\end{itemize}

\section{RELATED WORKS}
\subsection{Feature-based LiDAR-Camera Alignment Methods}

With the emergence of non-repetitive LiDAR, it is possible to obtain increasingly denser point clouds over time at stationary localizations, which can acquire more precise and distinct edge features from the environment, thereby facilitating the development of targetless LiDAR-camera calibration methods. In certain studies, \cite{yuan2021pixel, liu2022targetless} have proposed leveraging inherent 3D and 2D edges extracted from LiDAR maps and camera images, and \cite{miao2023coarse} designs an edge alignment method based on the edge feature of LiDAR intensity images. Extrinsic parameters are optimized by minimizing the distance between corresponding edge features in both sensors. Subsequently, each point cloud is transformed into the camera frame and projected onto the image plane to obtain corresponding RGB information. However, the colored point clouds are the indirect outcome of these methods. All these methods are designed for LiDAR-camera extrinsic calibration, which works only in edges' richness in natural scenes.

In addition, CP+ \cite{cui2022cp+} introduces a series of preprocessing operations, including removing blurry images and dynamic objects and selecting an appropriate region of interest (ROI) from the point cloud to address misalignment issues. However, feature mismatches and the absence of global constraints make these feature-based LiDAR-camera alignment methods unreliable in unstructured and textureless environments \cite{lin2022r}.
%

\subsection{Motion-based LiDAR-Camera State Estimation Solutions}

Paintcloud \cite{vechersky2018colourising} firstly proposes an offline method for point cloud colorization in a plug-and-play manner, and it can be seamlessly integrated with any LiDAR Odometry (LO) system \cite{zhang2014loam,shan2018lego}. However, the camera poses used in this method are calculated by linear interpolation with LiDAR trajectory, which leads to poor colorization results. 
By simply combining with the recent LiDAR-Visual-Inertial Odometry (LVIO) systems, the colorization result can be improved. LVI-SAM \cite{shan2021lvi} employs a factor graph as a tightly-coupled smoothing and mapping framework for sensor fusion. R2LIVE \cite{lin2021r} and R3LIVE \cite{lin2022r} effectively integrate data from LiDAR-Inertial-Visual sensors by minimizing the feature re-projection error and directly optimizing the frame-to-map photometric error, respectively. In addition, V-LOAM \cite{zhang2015visual} and DV-LOAM \cite{wang2021dv} fuse LiDAR-inertial-visual sensors at a level of loosely-coupled, in which the state is not jointly optimized with all sensor measurements.
The aforementioned Fusion-SLAM systems are specifically designed for localization and mapping applications, with some of them having already demonstrated available dense colored point clouds \cite{lin2022r, xu2021fast}. However, the cost function of these systems accounts for frame-to-map photometric residual without any global mapping consistency constraint consideration. 
Moreover, these SLAM-based online mapping techniques are unsuitable for accommodating customized high-resolution cameras in order to achieve enhanced colorization outcomes.

We draw inspiration from the photometric consistency mapping approach for mesh colorization \cite{zhou2014color}. However, this method is primarily designed for mesh representative reconstructions, limiting its applicability and robustness when extended to point clouds due to noise on the point cloud surface. We improve upon this by proposing a point cloud co-visibility estimation approach. Our algorithm is highly suitable for 360-degree cameras as it allows us to bypass the non-differentiable changes in the visibility relationship between the camera and point cloud during optimization. By seamlessly integrating high-resolution independent devices (360-degree cameras) into a mobile mapping system, our method ensures convenience and accuracy.

\begin{figure*}[ht!]
    \centering
    \includegraphics[width=0.9\textwidth]{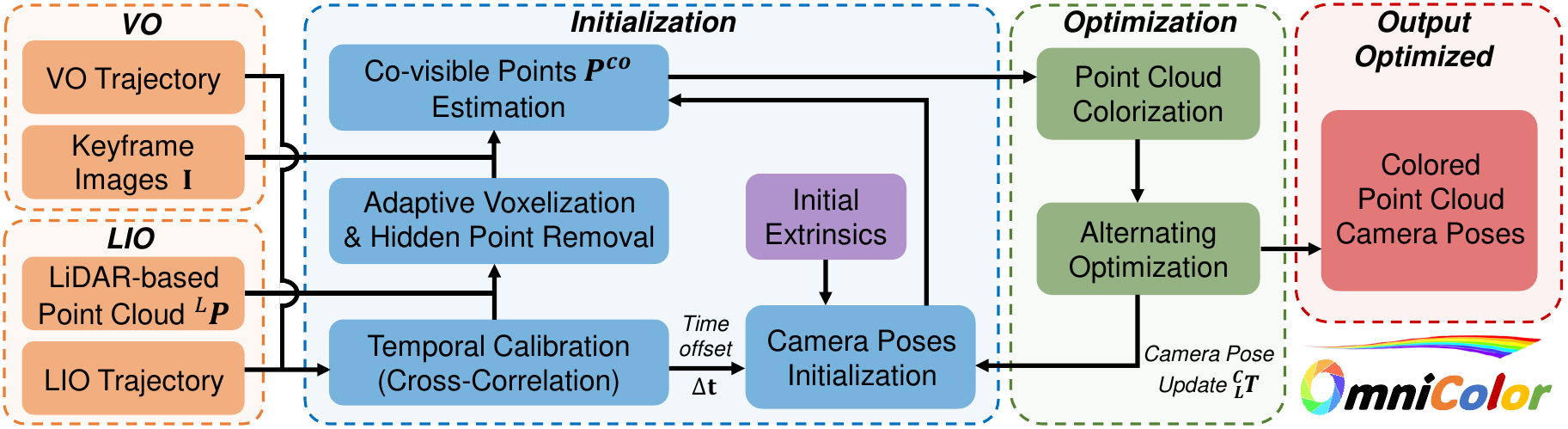}
    \vspace{-5pt}
    \caption{\small{Overview of our proposed system.
}}
    \label{Framework}
    \vspace{-15pt}
\end{figure*}


\vspace{0.5pt}
\section{METHODOLOGY}
\subsection{Overview} 


Fig.~\ref{Framework} presents an overview of our proposed methodology, in which the input is derived from two robustly established systems: the LiDAR-Inertial Odometry (LIO) system and the Visual Odometry (VO) system.
By employing LIO-based approaches \cite{xu2021fast,xu2022fast,bai2022faster}, we can acquire highly accurate LiDAR poses and enough corrected point cloud maps, facilitating the generation of a smooth and precise motion trajectory using high-frequency IMU data. Additionally, \cite{liu2021balm,liu2022efficient,yuan2022efficient} can be optionally employed as preprocessing steps to enhance the quality of the point cloud results. We denote $^{L}\textbf{P}$ as the final global point cloud input.
Meanwhile, we also implement a VO system \cite{sumikura2019openvslam,campos2021orb} to estimate the time offset $\Delta t$ between LiDAR and camera systems by comparing their motion signals through cross-correlation.

Given the handheld nature of the mobile mapping device, motion blur noticeably affects the quality of the input color images. In order to enhance the computed color point clouds, our pipeline autonomously chooses a subset of input images, utilizing a series of keyframes generated by the VO system as a basis for selection. Then, we evaluate and quantify the blurriness of each keyframe image using a no-reference metric \cite{ong2003no}. Subsequently, we add the frame with the lowest blurriness within each time segment $(t_-, t_+)$ after selecting the last keyframes. The selected image set is denoted as $\mathcal{I} = \{I_1,...I_n\}$. 
The coarse keyframe poses are calculated using linear interpolation based on the initial extrinsic parameters, temporal calibration result, and device trajectory. Here, we denote $n$ as the number of keyframes, $^{C}_{L} \mathcal{T} = \left\{^{C}_{L}\rm \mathbf{T}\it_1,...,^{C}_{L}\rm \mathbf{T}\it_n\right\} \in \it SE\rm (3)$ as the set of camera pose, and $^{C}_{L}\rm \mathbf{T}\it_i = \left(^{C}_{L}\rm \mathbf{R}\it_i, ^{C}_{L}\rm \mathbf{t}\it_i\right)$ as the $i$-th coarse camera pose $(i = 1,...,n)$. Note that the world frame, denoted as $W$, aligns with the LiDAR-based point cloud frame $L$, i.e., $\left(^{C}_{L}\rm \mathbf{R}\it_i, ^{C}_{L}\rm \mathbf{t}\it_i\right) = \left(^{C}_{W}\rm \mathbf{R}\it_i, ^{C}_{W}\rm \mathbf{t}\it_i\right)$. These color images are used in the subsequent stages of the pipeline. To provide a comprehensive explanation, the rest of the initialization stage's components (Section III-B) are individually described in detail. 
We begin Section III-B.1 with the method of point cloud adaptive voxelization and hidden points removal.
Subsequently, Section III-B.2 introduces the point cloud co-visibility estimation method. Section III-C demonstrates the effectiveness of our proposed approach in globally optimizing camera poses for all panorama images.


\subsection{Initialization}
\it{1)} Adaptive Voxelization in Hidden Points Removal: \rm 
The primary objective of this component is to identify the visible portion of the point cloud from a given viewpoint. \cite{vechersky2018colourising} involves applying the Hidden Point Removal (HPR) \cite{katz2015visibility} for this operation.
This operator involves two essential steps: point transformation and construction of a convex hull. However, the existing 3D convex hull algorithms are not well-suited to the parallel computing model of GPUs \cite{e2012efficient}, which makes them unsuitable for processing large-scale point clouds \cite{cui2022cp+}. 
To address this challenge, we enhance operational efficiency by leveraging the root nodes within the voxel map to predefine a maximum search distance, while the leaf nodes serve as a representation of the point cloud. This optimization significantly expedites the computation of the convex hull. However, it is essential to acknowledge that the utilization of fixed-resolution voxel maps entails a trade-off between segmentation time and accuracy. To better adapt to the environment, we have integrated an adaptive voxelization method \cite{liu2022targetless,liu2022efficient,yuan2022efficient} into our workflow to accelerate the process further.
\begin{figure}[ht]
    \centering
    \vspace{-8pt}
    \includegraphics[width=0.5\textwidth]{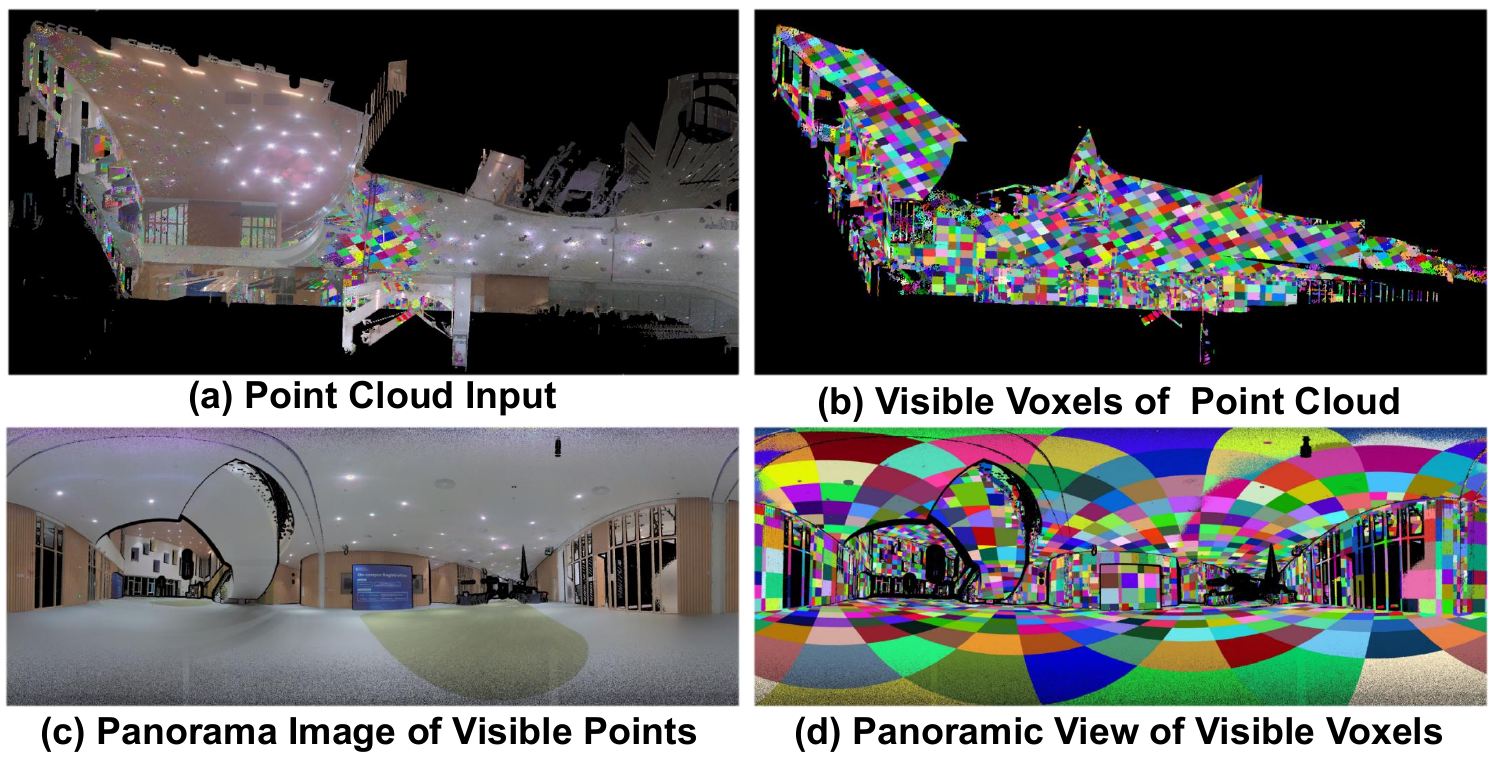}
    \vspace{-18pt}
    \caption{\small{Adaptive voxelization in Hidden Points Removal. The input point cloud is organized using an adaptive voxel-based data structure for accelerated processing. After adaptive voxelization, each voxel represents a flat surface, maintaining a consistent visible relationship.}}
    \label{HPR}
    \vspace{-5pt}
\end{figure}

\begin{figure}[ht]
    \centering
    \includegraphics[width=0.5\textwidth]{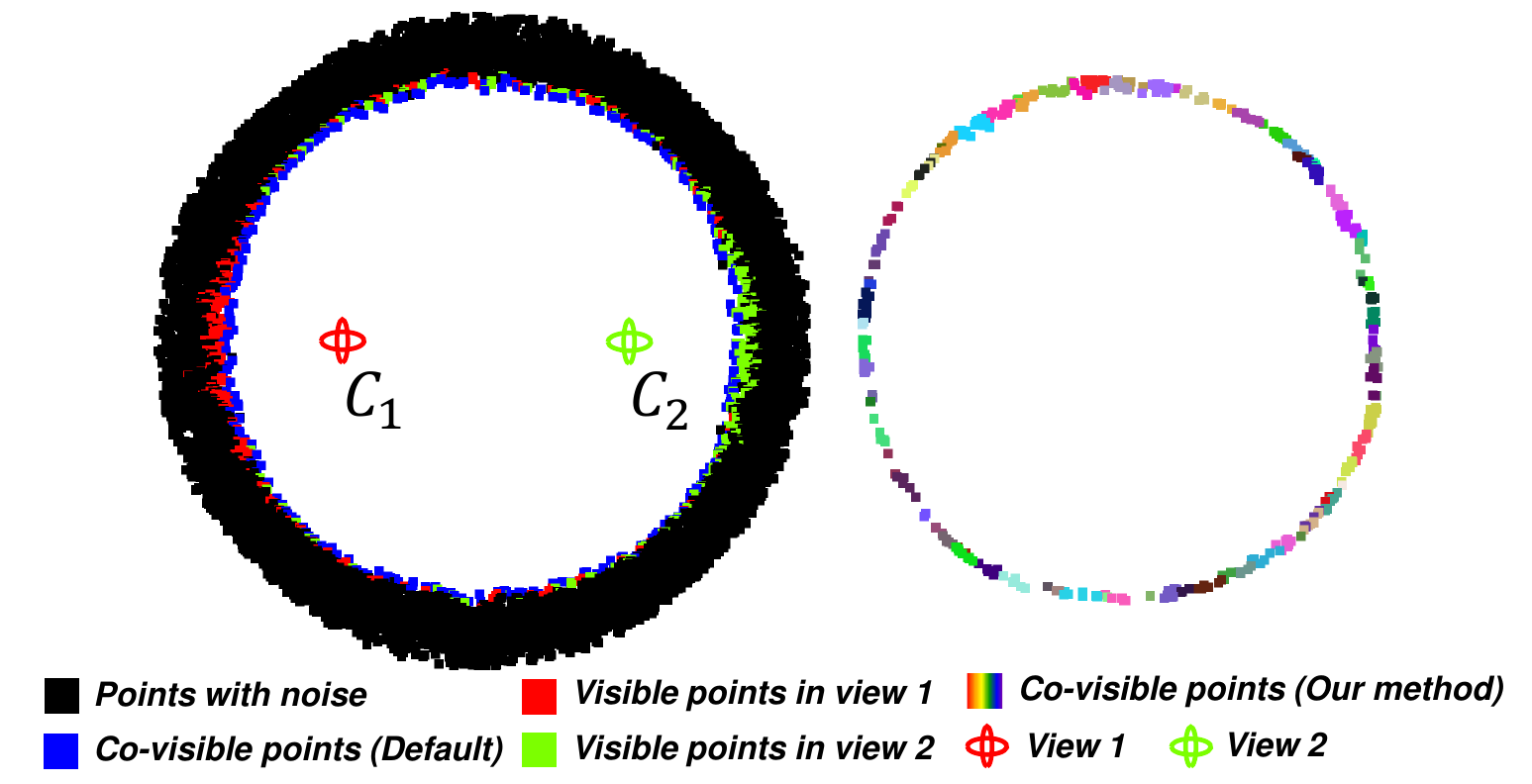}
    \caption{\small{An illustration of point cloud co-visibility estimation on a 2D circle simulated point cloud data.}}
    \label{omni}
    \vspace{-15pt}
\end{figure}

\it{2)} Point Cloud Co-visibility Estimation: \rm
We begin by outlining the pipeline for establishing a point cloud co-visibility estimation method, which aims to reduce the impact of point cloud surface noise on visibility relationships. To achieve this, we first subdivide the global point cloud into multiple voxels and assess the visibility of each point to determine the visible region for each keyframe $I_i$ based on the camera view. This operation is illustrated in Fig.~\ref{HPR} (a-d).
 The set of visible point clouds is denoted as $\textbf{P}^V = {P_1,..,P_n} \subset ^{L}\textbf{P}$, where $P_i = {\textit{p}_1^{idx_1},...,\textit{p}_m^{idx_m}}$ represents the set of visible points for keyframe $I_i$, $idx_i$ denotes the index of the voxel node that the point $p_i$ belongs to, and $m$ is the count of visible points. Next, we construct a co-visibility graph based on the co-visibility of the point cloud. If the count of visible points in two keyframes sharing the same voxel indexes exceeds a predefined threshold (e.g., half of the minimum count of visible points among these keyframes), an edge is established between them. The co-visible points are added to each keyframe's set of visible points, denoted as $P_i \rightarrow P_i^+$, where $P_i^+$ represents the updated visible points set for each keyframe. The set of co-visible points is denoted as $\textbf{\textit{P}}^{co} = \left\{P_i^+ \cap P_j^+ \right\}_{i \neq j}$. As shown in Fig.~\ref{omni}, we provide an example of the method's results mentioned above in a 2D space, offering a quantitative comparison between our approach and the default intersection method.

\subsection{Camera Pose Optimization}
\begin{figure}[ht]
    \centering
    \includegraphics[width=0.45\textwidth]{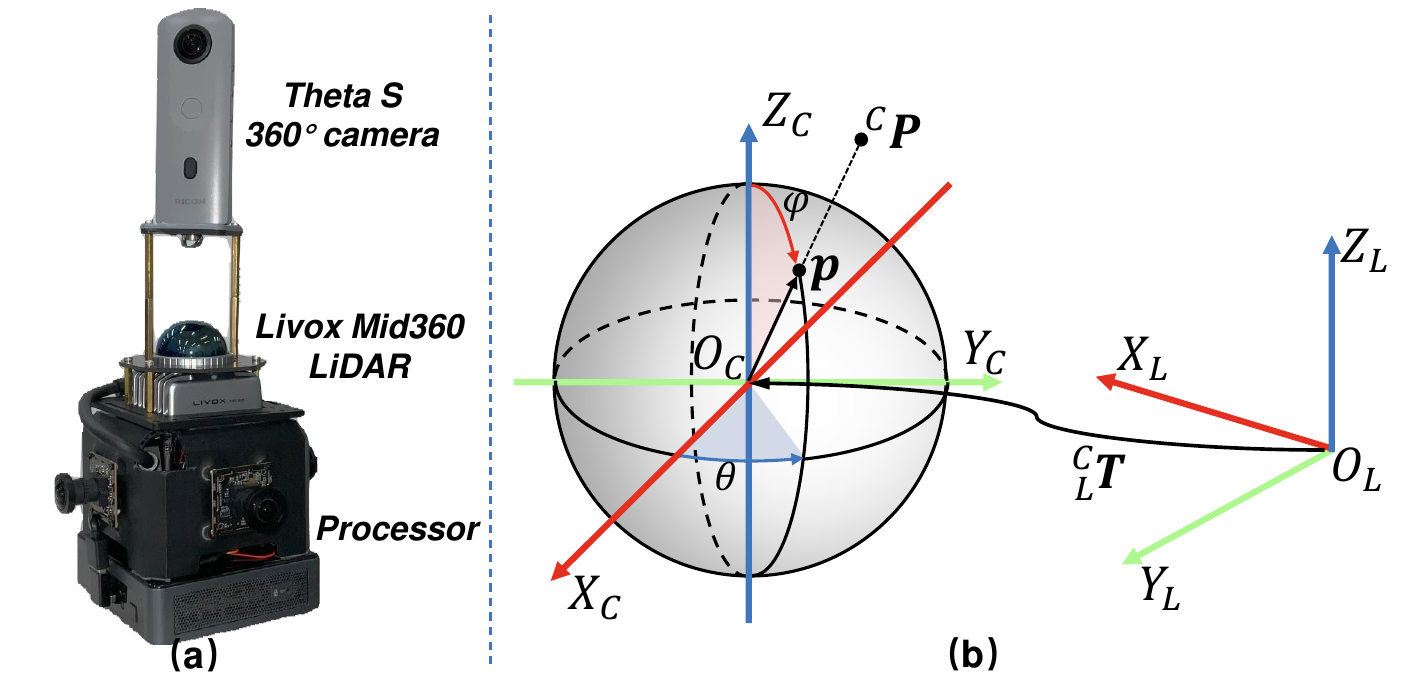}
    \vspace{-7pt}
    \caption{\small{(a) Our mobile mapping system and (b) LiDAR-camera coordinate system.}}
    \label{coord}
    \vspace{-8pt}
\end{figure}

\it{1)} Loss Function: \rm
Given the co-visible points set $\textbf{\textit{P}}^{co} = \left\{P_i^+ \cap P_j^+ \right\}_{i \neq j}$ and a sequence of keyframes $\mathcal{I}$, our objective is to find the optimal camera poses $^{C}_{L} \mathcal{T}$ at which panorama images are taken, where $L$ and $C$ are the LiDAR-based point cloud and camera frames, respectively. $^{C}_{L}\rm \mathbf{T}\it_i \in SE(3)$ denotes the extrinsic parameters from the LiDAR-based point cloud frame to the camera frame. Specifically, Fig.~\ref{coord} demonstrates the  LiDAR-based point cloud and camera coordinate systems, as well as the projection function $\Pi(\cdot): \mathbf{R}^{3}\rightarrow\mathbf{R}^{2}$ that maps a 3D point $^{C}\textbf{{p}}^{co} = \it({x},{y},{z})$ to a point $\textbf{{p}} = ({{u}}, {{v}}) = (\mathbf{\varphi},\mathbf{\theta})$ in the panorama images' coordinate frame, where $(\textbf{\textit{u}}, \textbf{\textit{v}}) \in[0, H)\times[0, W)$ and $(\varphi,\theta) \in (\frac{-\pi}{2},\frac{\pi}{2})\times[-\pi,\pi]$ denote the coordinates within the image plane and sphere space. $^{C}\textbf{{p}}^{co} = ^{C}_{L}\rm \mathbf{T}\it_i \textbf{{p}}^{co}$ project the point cloud from point cloud frame into the camera frame. This function could be explicitly written as follows:

\vspace{-5pt}
\begin{equation}
\begin{aligned}
\Pi(x,y,z) 
&= \left(\frac{H}{\pi}\arctan(\varphi), \frac{W}{2\pi}\arctan(\theta)\right), \\
(\varphi,\theta) 
&= \left(\frac{z}{\sqrt{x^2+y^2}}, \ \frac{y}{x}\right).
\end{aligned}
\end{equation}

Furthermore, let $\Gamma_i(\textbf{{p}}, I_i)$ indicate the function that maps the 2D coordinates $\textbf{{p}}$ to pixel values from the keyframes $ I_i\in \mathcal{I}_{\textbf{{p}}}$, where $\mathcal{I}_{\textbf{{p}}}$ denotes the set of images associated with $\textbf{p}$. 
Under this setup, $\Gamma_i\left(\Pi\left(^{C}_{L}\rm \mathbf{T}\it_i\textbf{{p}}^{co}\right), I_i\right)$ be the color at the image coordinates of the projection of $\textbf{\textit{p}}^{co}$ onto $I_i$, given an extrinsic matrix $^{C}_{L}\rm \mathbf{T}\it_i$. 
We want to maximize the agreement within $\left\{\Gamma_i\left(\Pi\left(^{C}_{L}\rm \mathbf{T}\it_i\textbf{{p}}^{co}\right),I_i\right)\right\}_{I_i\in \mathcal{I}_{\textbf{p}}}$ for each co-visible point $\textbf{{p}}^{co}$. 
To this end, we introduce an auxiliary variable $C(\textbf{{p}}^{co})$ to represent the color of $\textbf{{p}}^{co}$. 
If the co-visible point $\textbf{{p}}^{co}$ is perfectly aligned with the keyframes $\textbf{I}$, one could except the projected pixel values' set $\left\{\Gamma_i\left(\Pi\left(^{C}_{L}\rm \mathbf{T}\it_i\textbf{{p}}^{co}\right),I_i\right)\right\}_{I_i\in \mathcal{I}_{\textbf{p}}}$ to be very close to the point color values $C(\textbf{{p}}^{co})$. 
Our goal is to optimize the set of camera pose $^{C}_{L} \mathcal{T}$, with the set of the auxiliary variable $\textbf{C} = \{C(\textbf{{p}}^{co})\}$, where the objective is to minimize the discrepancy between $\left\{\Gamma_i\left(\Pi\left(^{C}_{L}\rm \mathbf{T}\it_i\textbf{{p}}^{co}\right),I_i\right)\right\}_{I_i\in \mathcal{I}_{\textbf{p}}}$ and $C(\textbf{{p}}^{co})$. 
Let $\textbf{P}^{co}_i = P_i^+\cap \{ P_j^+ \}_{i \neq j}$, where $\textbf{P}^{co}_i$ denotes the co-visible points set from keyframes $I_i$.
This can be formulated as follows:

\vspace{-10pt}
\begin{equation}
E(\textbf{C},^{C}_{L} \mathcal{T}) =\sum_i \sum_{\textbf{p}^{co}\in \textbf{P}^{co}_i}\|\Gamma_i(\Pi\left(^{C}_{L}\rm \mathbf{T}\it_i\textbf{{p}}^{co}\right))-C(\textbf{{p}}^{co})\| \rm_2.
   \label{eq3}
\end{equation}

The loss function from Equation~\ref{eq3}. is the point cloud-centric sampling loss, which is evaluated at the projected location of every co-visible point in the point cloud.
This sampling loss has several advantages. First, It fairly incorporates all points in the point cloud, thus making it overcome the severe visual distortion of omnidirectional images. In addition, the $\textbf{P}^{co}_i$ for each image $I_i$ is definitely stable for 360-degree cameras. Thus, it allows us to bypass the non-differentiable changes in the visibility relationship between the camera and point cloud during optimization, significantly reducing computational costs.

\it{2)} Alternating Optimization: \rm
We have implemented an alternating optimization scheme, inspired by \cite{zhou2014color}, to minimize the loss function. The basic idea is to alternate optimizing between $^{C}_{L}\mathcal{T}$ and $\textbf{C}$. When optimizing $\textbf{C}$, $^{C}_{L}\mathcal{T}$ is kept fixed, and vice versa.
To initialize the camera poses when $^{C}_{L} \mathcal{T}$ is fixed, we use the extrinsic calibration between the camera and the LiDAR Odometry. We then colorize the point cloud based on the initial coarse camera poses and keyframe images. However, due to slight inaccuracies in the point cloud map, illumination changes from different camera views, and coarse camera poses, colorization errors are inevitable. To mitigate this problem, we use a form of robust average.
Each point in the point cloud has a set of candidate colors, and the number of the candidate colors is denoted as $\textbf{k}$.
Then, the nonlinear problem Equation~\ref{eq3} is transformed into a linear least-squares problem with a closed-form solution.

\vspace{-5pt}
\begin{equation}
C(\textbf{{p}}^{co}) = \frac{1}{\textbf{k}_{\textbf{p}^{co}}} \sum_{I_i \in \mathcal{I}_{\textbf{p}^{co}}} \Gamma_i(\Pi\left(^{C}_{L}\rm \mathbf{T}\it_i\textbf{{p}}^{co}\right)).
   \label{eq4}
\end{equation}

When $\textbf{C}$ is fixed, the loss function Equation~\ref{eq3}. decomposes into an independent loss for each $^{C}_{L}\rm \mathbf{T}\it_i$:

\vspace{-5pt}
\begin{equation}
E_i(\mathbf{T}) = \sum_{\textbf{p}^{co}\in \textbf{P}^{co}_i}\left\|\Gamma_i(\Pi\left(^{C}_{L}\rm \mathbf{T}\it_i\textbf{{p}}^{co}\right))-C(\textbf{{p}}^{co})\right\| \rm_2.
   \label{eq5}
\end{equation}

\begin{figure}[ht]
    \centering
    \includegraphics[width=0.45\textwidth]{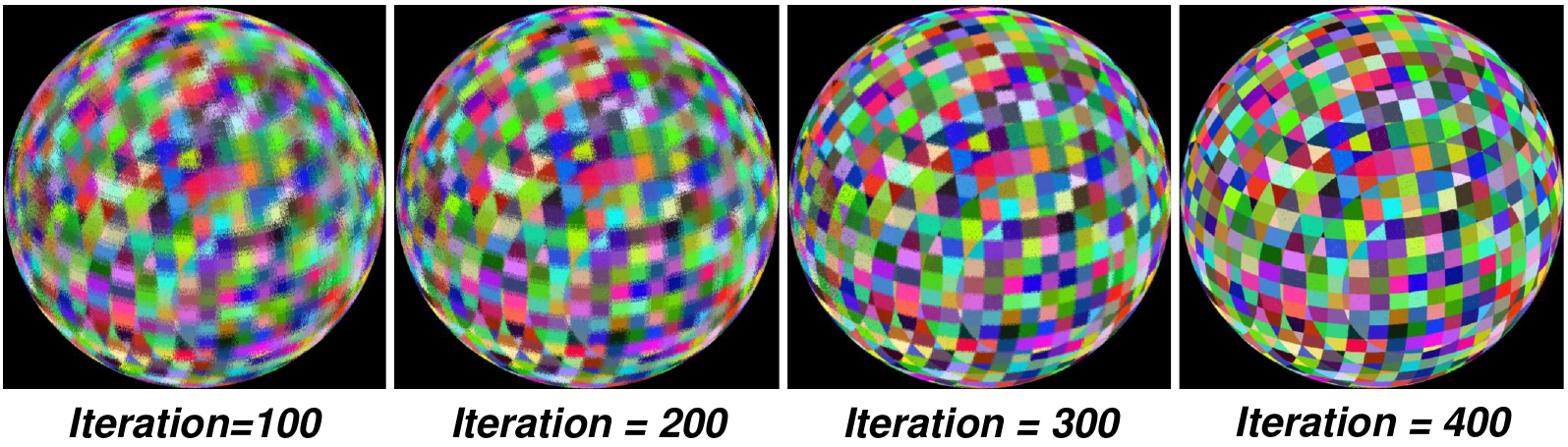}
    \vspace{-7pt}
    \caption{\small{The colorization result during optimization process. Our method is capable of processing sphere simulation point clouds, which inherently lack distinct geometric features.}}
    \label{noise}
    \vspace{-10pt}
\end{figure}

The core part of OmniColor consists of a simple gradient descent on the sampling loss Equation~\ref{eq5}, which is computationally efficient. Meanwhile, the loss function can effectively handle the comprehensive analysis of the 360-degree image and demonstrate robustness against visual distortion. Fig.~\ref{noise} shows an example of our loss function optimization process on the simulation dataset.

\vspace{-5pt}

\section{EXPERIMENTS}
\begin{figure*}[ht!]
    \centering
    \includegraphics[width=1.0\textwidth]{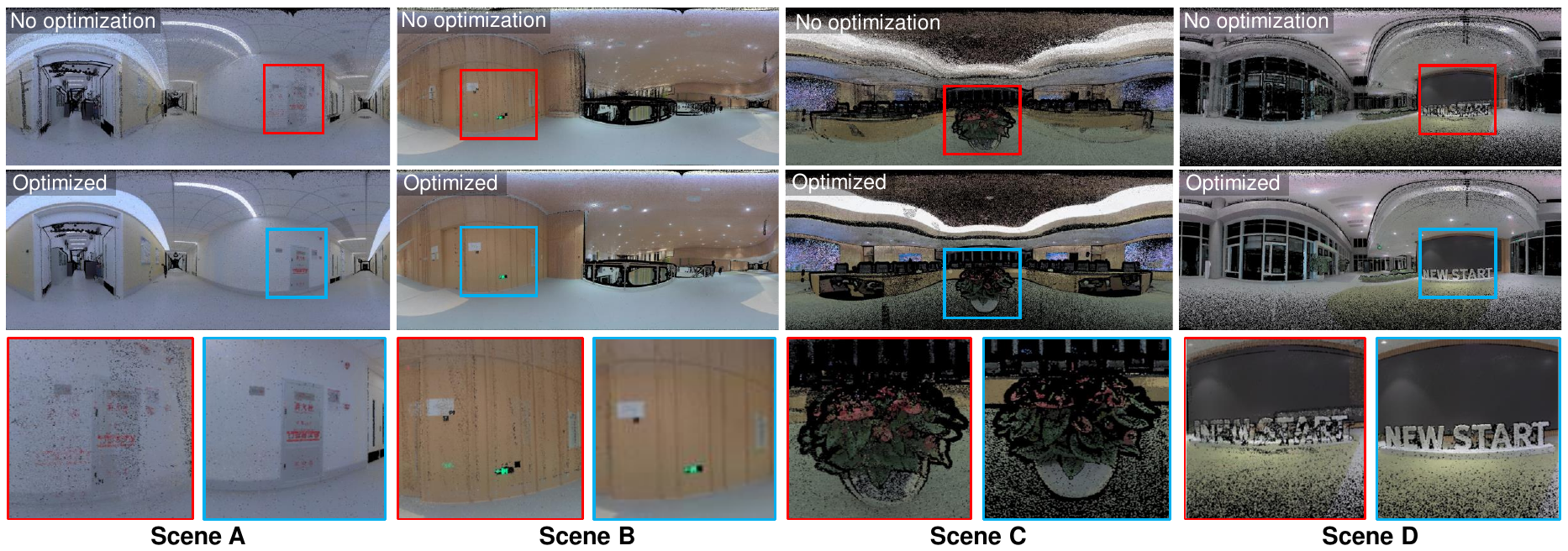}
    \vspace{-20pt}
    \caption{The rendered images of the colorization result of the HKUST Guangzhou campus dataset (Scene A-D). Left-red: The colored point with noise before optimization. Right-blue: The colored point cloud optimized by our method. }
    \label{result}
    \vspace{-1pt}
\end{figure*}

To evaluate the performance of the proposed method, we compare it against other state-of-the-art approaches using two categories of datasets. The first includes data collected from a mobile mapping device, specifically the Livox Mid360 solid-state LiDAR for high-resolution point cloud measurements and a RICOH Theta S 360-degree camera for qualitative analysis (see Fig.~\ref{result}). The second category comprises data captured from a stationary LiDAR Scanner named BKL360 within the HKUST Guangzhou campus premises, including four scenes with over 10 station frames from various positions and orientations. Additionally, we utilize the publicly available Omniscenes dataset \cite{kim2021piccolo}, consisting of panorama image sequences with ground truth camera poses and scene 3D point clouds, for comparison with the SfM-based method \cite{lee2021large}. Finally, a simulated dataset is employed for an ablation study on our method regarding point cloud co-visibility estimation.

To quantitatively compare our optimized parameters with ground-truth values in terms of rotation error (unit: degree) and translation error (unit: centimeters), random rotational and translational noise is added to the original camera views before running the optimization algorithm. The initial rotation and translation errors are computed as $5.0^\circ/10cm$. The following presentation of experimental results aims to directly demonstrate the efficacy of our method through quantitative and qualitative analysis.

\begin{table}[ht]
\centering
\setlength{\tabcolsep}{1.3mm}
\vspace{-5pt}
\begin{tabular}{@{}clllll@{}}
\multicolumn{6}{c}{TABLE I}
\\
\multicolumn{6}{c}{\begin{tabular}[c]{@{}c@{}}ROTATION ERRORS (DEGREES) AND TRANSLATION ERRORS\\ (CENTIMETERS) ON HKUST GUANGZHOU CAMPUS DATASET\end{tabular}}
\\ \midrule
\multicolumn{1}{l}{{\color[HTML]{000000} \textit{\textbf{}}}}  & {\color[HTML]{000000} Scene 1} & {\color[HTML]{000000} Scene 2} & {\color[HTML]{000000} Scene 3} & Scene 4 & Average 
\\ \midrule
{\color[HTML]{000000} \begin{tabular}[c]{@{}c@{}}Depth Edge-\\ Based {}\cite{yuan2021pixel}{}\end{tabular}} & {\color[HTML]{000000} 0.89/5.45} & {\color[HTML]{000000} 1.52/6.75} & {\color[HTML]{000000} 0.76/4.78}  & {\color[HTML]{000000} 3.67/5.81} & 1.71/5.70  \\
{\color[HTML]{000000} \begin{tabular}[c]{@{}c@{}}Intensity Edge-\\ Based {}\cite{miao2023coarse}{}\end{tabular}} & {\color[HTML]{000000} 0.82/4.53} & {\color[HTML]{000000} 0.96/3.67} & {\color[HTML]{000000} 1.98/6.83} & {\color[HTML]{000000} 4.37/7.21} & 2.03/5.56  \\
{\color[HTML]{000000} \textbf{Ours}} & {\color[HTML]{000000} \textbf{0.05/3.83}} & {\color[HTML]{000000} \textbf{0.03/2.56}} & {\color[HTML]{000000} \textbf{0.04/2.48}} & {\color[HTML]{000000} \textbf{0.07/3.35}} & \textbf{0.475/3.06}
\\ \midrule
\multicolumn{6}{c}{*The initial rotation and translation errors are 5° /10 cm.}
\label{Table1}
\vspace{-8pt}
\end{tabular}
\end{table}

\begin{table*}[ht]
\setlength{\tabcolsep}{2.3mm}
\vspace{1pt}
\centering
\begin{tabular}{cccccccccccccccccccccc}
\multicolumn{9}{c}{TABLE II}
\\
\multicolumn{9}{c}{ROTATION ERRORS (DEGREES) AND TRANSLATION ERRORS (CENTIMETERS) ON OMNISCENES DATASET}
\\ \midrule
{\color[HTML]{000000} \textit{\textbf{}}}            & {\color[HTML]{000000} PyebaekRoom}        & {\color[HTML]{000000} Room 1}             & {\color[HTML]{000000} Room 2}             & Room 3                                    & Room 4             & Room 5             & WeddingHall 1      & Average \\ 
\midrule
{\color[HTML]{000000} Prior Pose-based SfM \cite{lee2021large}} & {\color[HTML]{000000} 0.26/2.75}          & {\color[HTML]{000000} 0.34/4.25}          & {\color[HTML]{000000} 0.26/3.46}          & {\color[HTML]{000000} 0.35/3.57}          & 0.43/\textbf{3.42}          & 0.35/4.53          & \textbf{0.21/3.26} & 0.31/3.61\\ 
{\color[HTML]{000000} \textbf{Ours}}                 & {\color[HTML]{000000} \textbf{0.12/2.54}} & {\color[HTML]{000000} \textbf{0.24/3.42}} & {\color[HTML]{000000} \textbf{0.18/2.96}} & {\color[HTML]{000000} \textbf{0.34/3.42}} & \textbf{0.38}/4.28 & \textbf{0.28/3.53} & 0.24/4.75          & \multicolumn{1}{c}{\textbf{0.25/3.56}} \\
\midrule
\multicolumn{9}{c}{*The initial rotation and translation errors are 5° /10 cm.}
\end{tabular}
\label{Table2}
\end{table*}

\setlength{\parindent}{0pt}
\it{A.} Quantitative Results \rm
\vspace{2pt}
\setlength{\parindent}{10pt}

Our method is specifically tailored for the utilization of a 360-degree camera. To facilitate comparison with other methods, we rectify the panorama image into a pinhole image encompassing FOV of 160 degrees and subdivide the input point cloud into multiple local maps centered around coarse camera views. 
Specifically, we compare our methods with two types of edge feature-based extrinsic calibration approaches: depth-continuous edges \cite{yuan2021pixel} in the point cloud and edge features extracted from LiDAR intensity images \cite{miao2023coarse}, which align the local maps with color images through extrinsic calibration.
We have fine-tuned the parameters of each method to achieve the best performance through the authors' efforts. Consistent parameter settings are applied across all scenes for each method.

The experimental results are summarized in Table I. Our method outperforms other approaches in terms of accuracy across all scenes in the HKUST Guangzhou campus dataset. In contrast to edge feature-based methods, which exhibit sensitivity to the environment and encounter failures in several local maps, our approach demonstrates robustness in handling diverse mapping scenes. We acknowledge that there were failure cases with the edge feature-based method; therefore, for calculating the final errors of edge feature-based methods presented in Table I, we only evaluate successfully aligned local maps. However, even within these successful scenes, our method consistently achieves superior accuracy.
The improved performance may be attributed to two reasons. Firstly, the color space of the omnidirectional camera is more sensitive than the edge features proposed in \cite{yuan2021pixel,miao2023coarse}. In our specific scenario, the initial noise is minimal, resulting in coarse camera poses that are close to the global minimum solution where feature-based methods' gradient approaches zero. However, our direct method still possesses a clear optimized direction leading to more precise outcomes. Additionally, it's challenging to identify reliable correlation features from various modalities, which can cause edge feature mismatches and degrade alignment accuracy.

We further evaluate with the SfM method \cite{lee2021large}, which jointly optimizes image poses, feature points, and calibration parameters using LiDAR-SLAM as priors. We conduct experiments on the Omniscenes dataset. This dataset contains panorama videos with ground truth poses and 3D point clouds, which can be regarded as the data captured by a mobile mapping system. For a fair comparison, we convert the panorama image to six cube pinhole images and initialize the coarse camera poses by randomly perturbing the ground-truth values ($5.0^\circ$ for rotation; $10cm$ for translation). In the optimization processing, we add rigid constraints on the six cube pinhole images derived from the same panorama image. The results are shown in Table II. We notice that the accuracy of our approach is close to the joint SfM method \cite{lee2021large}. Although achieving approximate accuracy, ours is a kind of direct method that does not need any feature extraction and matching process that can significantly reduce computational costs. In addition, our optimization scheme makes it easy to exploit modern GPU power to update all poses and point cloud colorization results in parallel.
\begin{figure}[ht!]
    \centering
    \vspace{-5pt}
    \includegraphics[width=0.45\textwidth]{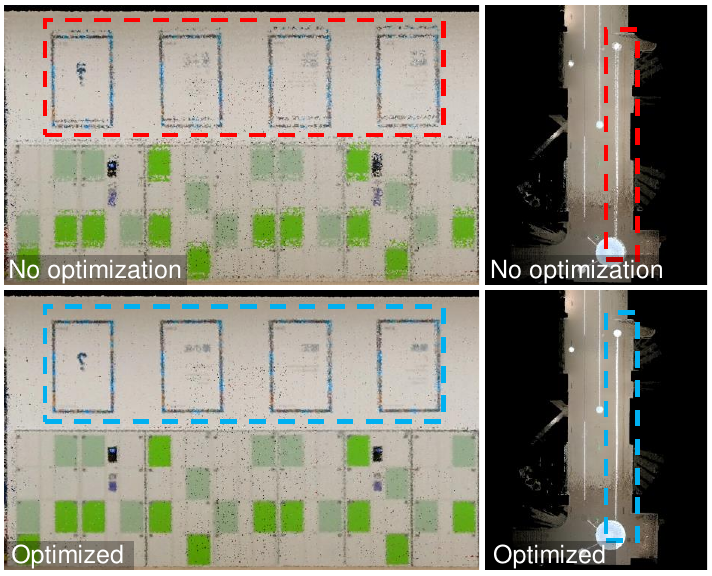}
    \vspace{-5pt}
    \caption{The colored point cloud reconstructed by our mobile mapping device. Top: Colored based on interpolated camera poses. Bottom: colored based on optimized camera poses}
    \label{mobile result}
    \vspace{-20 pt}
\end{figure}

\vspace{3pt}
\setlength{\parindent}{0pt}
\it{B.} Qualitative Results \rm
\setlength{\parindent}{10pt}

In addition to the qualitative comparison, we have specifically chosen various failure scenarios involving feature-based methods from different sequences within the HKUST Guangzhou campus dataset. These scenarios highlight the consistent high-quality results achievable with our method across all scenes.
Fig.~\ref{result} illustrates how our method significantly enhances point cloud colorization outcomes. In the case of the mobile mapping dataset, we conducted a comparative analysis against \cite{vechersky2018colourising}, as shown in  Fig.~\ref{mobile result}. The results clearly indicate the substantial improvement in colorization achieved by our method.

\vspace{3pt}
\setlength{\parindent}{0pt}
\it{C.} Ablation Study on Point Cloud Co-visibility Estimation \rm
\setlength{\parindent}{10pt}

\begin{figure}[ht]
    \centering
    \vspace{-12pt}
    \includegraphics[width=0.5\textwidth]{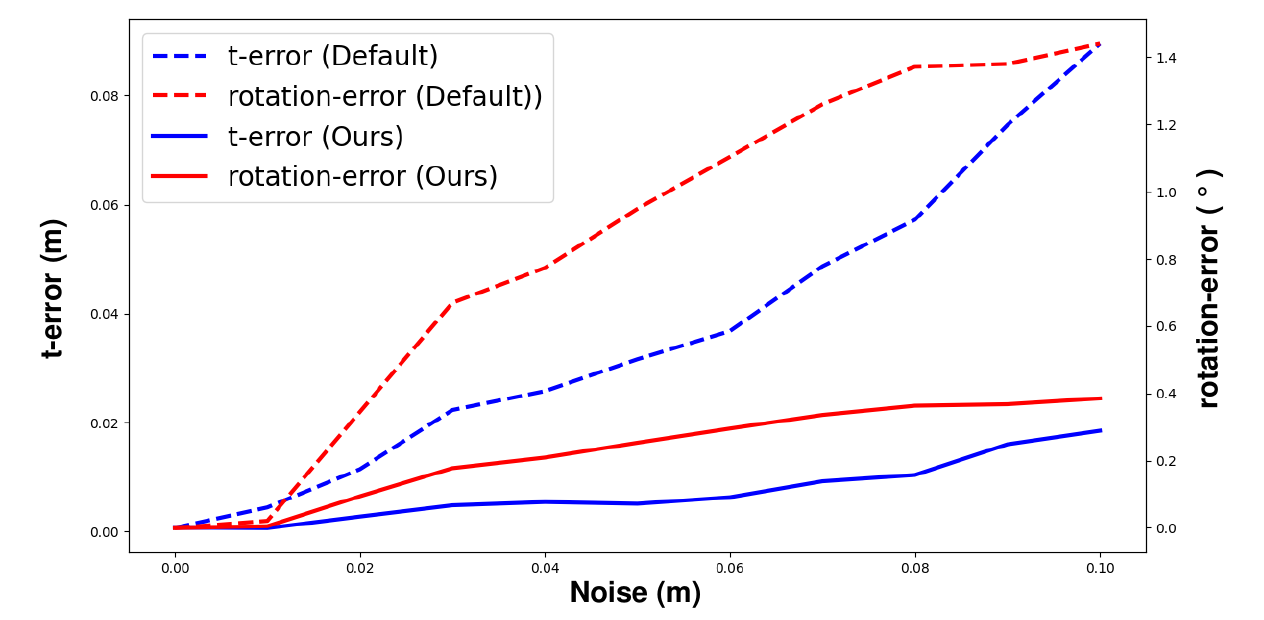}
    \vspace{-10pt}
    \caption{Ablation study about the points co-visibility estimation method for camera pose optimization.}
    \label{ablation_study}
    \vspace{-10pt}
\end{figure}

We perform this experiment using a simulated dataset that we generated, consisting of points within a sphere with a radius of 10m. Fig.~\ref{noise} illustrates the progress of our colorization process during optimization. In this section, we conduct a quantitative analysis to assess the impact of point noise on the optimization results.
Fig.~\ref{ablation_study} introduces noise ranging from 1cm to 10cm. The results demonstrate that our method effectively mitigates the noise impact on the point cloud's surface, leading to more precise camera poses.

\section{CONCLUSIONS}
In this paper, we propose a novel and efficient algorithm (OmniColor) to colorize point clouds using an independent 360-degree camera. It takes advantage of the omnidirectional perception view of the camera and eliminates the impact of severe spherical distortion while maintaining computational efficiency.
In experiments conducted on both our proprietary dataset and the public dataset, OmniColor outperforms existing algorithms in both accuracy and stability compared to SOTA methods.
Moreover, OmniColor seamlessly integrates with any mobile mapping system, opening up a world of possibilities for its application in various domains, including virtual reality and robotics, where the need for clear and precise colored point cloud maps is paramount.


\addtolength{\textheight}{-2cm}   






\clearpage



\bibliographystyle{IEEEtran}
\bibliography{IEEEfull}

\end{document}